\documentclass[letterpaper, 10 pt, conference]{ieeeconf}  

\IEEEoverridecommandlockouts                              

\usepackage{graphics}
\usepackage{epsfig} 
\usepackage{mathptmx} 
\usepackage{times}
\usepackage{amsmath,amssymb,amsfonts,bm}
\usepackage{graphicx}
\usepackage{algorithm,algorithmic}
\usepackage{flushend}
\usepackage{booktabs,threeparttable,multirow, caption}
\usepackage{color} 
\usepackage{cuted}
\usepackage{array}
\usepackage[sorting=none, style=ieee]{biblatex}
\def\BibTeX{{\rm B\kern-.05em{\sc i\kern-.025em b}\kern-.08em
    T\kern-.1667em\lower.7ex\hbox{E}\kern-.125emX}}
\begin{document}

\title{EndoOmni: Zero-Shot Cross-Dataset Depth Estimation in Endoscopy by Robust Self-Learning from Noisy Labels\\

}

\author{Qingyao Tian, Zhen Chen, Huai Liao, Xinyan Huang, Lujie Li, Sebastien Ourselin, Hongbin Liu
\thanks{
Qingyao Tian is with Institute of Automation, Chinese Academy of Sciences, Beijing, China, and also with the School of Artificial Intelligence, University of Chinese Academy of Sciences, Beijing, China.
}
\thanks{
Zhen Chen is with Centre of AI and Robotics, Hong Kong Institute of Science\& Innovation, Chinese Academy of Sciences, HK, China.
}
\thanks{
Huai Liao, Xin-yan Huang and Lujie Li are with The First Affiliated Hospital of Sun Yat-sen University, Guangzhou, Guangdong, China.
}
\thanks{
Sebastien Ourselin is with the School of Engineering and Imaging Sciences, King’s College London, UK.
}
\thanks{
Corresponding author: Hongbin Liu is with Institute of Automation, Chinese Academy of Sciences, and with Centre of AI and Robotics, Hong Kong Institute of Science\& Innovation, Chinese Academy of Sciences. He is also affiliated with the School of Engineering and Imaging Sciences, King’s College London, UK. (e-mail: liuhongbin@ia.ac.cn).
}}

\maketitle
\begin{strip}
\begin{minipage}{\textwidth}\centering
\vspace{-100pt}
\includegraphics[width=\textwidth]{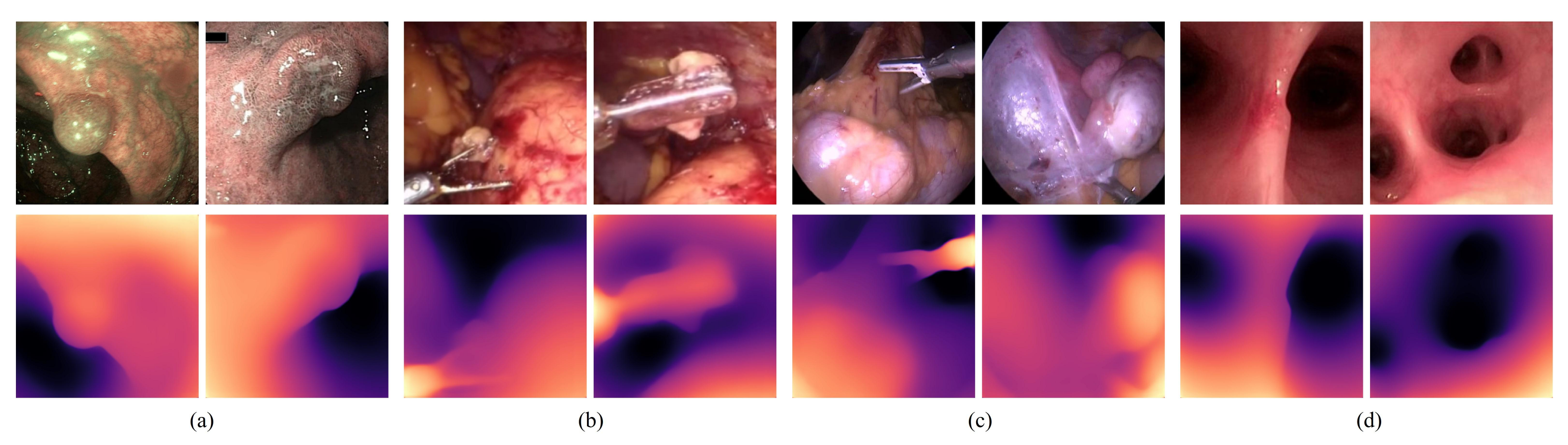}
\captionof{figure}{EndoOmni demonstrates exceptional zero-shot performance across various unseen endoscopy datasets: \textbf{(a)} the OBR dataset \cite{ye2016online}, \textbf{(b)} the da Vinci surgical dataset \cite{ye2017self}, \textbf{(c)} the Heico dataset \cite{maier2021heidelberg}, and \textbf{(d)} our own bronchoscopy data collected from porcine models. The scenes encompass a range of environments, including tubular structures, complex surgical tools and intricate lumen hierarchy.}

\label{figurelabel}
\end{minipage}
\end{strip}

\begin{abstract}
Single-image depth estimation is essential for endoscopy tasks such as localization, reconstruction, and augmented reality. Most existing methods in surgical scenes focus on in-domain depth estimation, limiting their real-world applicability. This constraint stems from the scarcity and inferior labeling quality of medical data for training. In this work, we present EndoOmni, the first foundation model for zero-shot cross-domain depth estimation for endoscopy. To harness the potential of diverse training data, we refine the advanced self-learning paradigm that employs a teacher model to generate pseudo-labels, guiding a student model trained on large-scale labeled and unlabeled data. To address training disturbance caused by inherent noise in depth labels, we propose a robust training framework that leverages both depth labels and estimated confidence from the teacher model to jointly guide the student model training. Moreover, we propose a weighted scale-and-shift invariant loss to adaptively adjust learning weights based on label confidence, thus imposing learning bias towards cleaner label pixels while reducing the influence of highly noisy pixels. Experiments on zero-shot relative depth estimation show that our EndoOmni improves state-of-the-art methods in medical imaging for 33\% and existing foundation models for 34\% in terms of absolute relative error on specific datasets. Furthermore, our model provides strong initialization for fine-tuning metric depth estimation, maintaining superior performance in both in-domain and out-of-domain scenarios. The source code is publicly available at {\url{https://github.com/TianCuteQY/EndoOmni}}.
\end{abstract}


\newcolumntype{C}[1]{>{\centering\arraybackslash}p{#1}}

\section{Introduction}
Depth estimation is a critical task for numerous applications such as 3D reconstruction, localization and augmented reality visualization \cite{ju2023dg,kastner20203d,huang2024photo}. In the realm of single-image depth estimation (SDE) for natural images, significant strides have been made toward developing generalizable foundation models across various datasets \cite{ranftl2020towards,ranftl2021vision,yang2024depth,yang2024depth2}. Their efficacy stems from extensive training on large-scale datasets, enabling versatile representation learning. 

However, the SDE in endoscopic imagery remains largely domain-specific, suffering from poor generalization and limited real-world application. Despite some efforts to adapt foundation models trained on natural imagery for medical depth estimation, these are limited in scope and primarily experimental \cite{cui2024surgical,cui2024endodac,han2024depth}. The disparity in research focus between natural and medical imaging can be attributed to two challenges inherent in medical datasets: \textbf{1) data scarcity} \cite{cui2020unified} and \textbf{2) inferior labeling quality} \cite{xue2022robust}. These factors result in models that are prone to overfitting and struggle to generalize to unseen data \cite{zhon2024robust}.

This study aims to develop a foundation model for endoscopic SDE by addressing these challenges, enabling robust zero-shot performance across diverse datasets while offering fine-tuning for state-of-the-art metric depth estimation.

Addressing the limitation of data scarcity, we have assembled the most extensive meta-dataset for medical image SDE, encompassing over 200,000 labeled frames from public and proprietary datasets and an additional 500,000 unlabeled frames (Table \ref{tab:datasets}) for model training and evaluation. To harness the potential of unlabeled data, we leverage a self-learning strategy that utilizes a pretrained teacher model to generate pseudo labels. This combined labeled and pseudo-labeled data is then used to train a student model that processes highly perturbed images. This self-learning framework enhances training data variety, promoting generalization by exposing the model to diverse features and patterns. However, since the teacher model predicts out-of-domain datasets, the resulting pseudo labels are often noisy, posing an additional challenge for robust training.

To handle noisy labels, previous SDE studies have introduced a trimming strategy for loss calculation, where outliers are detected and removed by trimming the largest residuals in each instance \cite{ranftl2020towards}. While this strategy contributes to training stability, it tends to neglect challenging image regions. The recent Depth Anything v2 \cite{yang2024depth2} also recognized the performance drop due to label noise and suggested using synthetic data and a stronger pretrained teacher model for training. However, these solutions are impractical in the medical imaging domain due to the lack of large synthetic datasets and the unavailability of strong pretrained models.

In this work, we pursue a refined self-learning framework and a robust training loss to guide network training by noisy labels and pseudo labels. We creatively extend the line of thinking in robust training on classification problems \cite{zhou2020robust,liu2020early,li2023disc} to the challenge of depth estimation. Through our empirical experiments, we observe that loss on clean data converges faster due to the inherent consistency of clean labels. Additionally, pseudo-labels tend to be more reliable when the teacher model predicts consistent results for the same instance across different augmentations. 
Based on this analysis, we propose estimating per-pixel label confidence according to the teacher model's learning behavior, and using both labels and confidence to jointly guide the training of the student model. Moreover, we have developed a robust SDE training loss that adjusts learning weights based on estimated label confidence. This approach not only mitigates training instability caused by label noise but also maintains the contribution of difficult regions in each instance, thereby enhancing the learning effectiveness of the SDE network.

In summary, the contributions of this work are as follows:

\begin{itemize}
    \item We present EndoOmni, the first foundation model for zero-shot cross-dataset depth estimation of endoscopic imaginary, by leveraging the potential of large-scale comprehensive endoscopy data.

    \item We propose a robust self-learning framework where a student model learns from a mix of labeled and unlabeled data, guided by the label confidence derived from analyzing the teacher model's learning behavior.

    \item We address the disturbance of noisy labels by proposing a weighted scale-and-shift-invariant training loss. This approach adjusts learning weights based on estimated label confidence, promoting learning bias towards clean labels and ensuring robust training.

    \item Our model exceeds existing methods in medical imaging and foundation models in zero-shot relative depth estimation with a large margin. Furthermore, it maintains superior performance in both in-domain and out-of-domain scenarios fine-tuned to metric depth estimation.
\end{itemize}

\section{Related Works}
Our main contribution is a foundation model for single-image depth estimation (SDE) in endoscopy by learning from noisy labels. Thereby, we review related works on cross dataset SDE, SDE in endoscopy and robust learning.

\vspace{0.3cm}
\noindent \textbf{Cross Dataset SDE.} Contrary to works that learn depth estimation model for a single dataset \cite{bhat2021adabins,bhat2022localbins,huynh2020guiding}, emerging research \cite{ranftl2020towards,ranftl2021vision,yang2024depth,piccinelli2024unidepth} has explored training SDE models or pretraining flagship models on mixing dataset. The pioneering work by MiDaS \cite{ranftl2020towards} introduced affine-invariant loss, allowing for scale-and-drift agnostic depth learning across datasets, thus mitigating the variance in depth distributions. Subsequent developments, such as DPT \cite{ranftl2021vision}, have leveraged more complex architectures like the vision transformer to further enhance model performance. Depth Anything \cite{yang2024depth} extended this paradigm by incorporating unlabeled data into the training process, which significantly boosted the performance of MDE models. While these models initially predict only relative depth, their ability to adapt to metric depth estimation through fine-tuning has demonstrated superior performance compared to models trained on isolated datasets.

\vspace{0.3cm}
\noindent \textbf{SDE in Medical Endoscopy.} SDE in endoscopic images aids in precise navigation, accurate lesion assessment, and improved surgical planning. Given the impracticality of using depth cameras or LiDAR in such settings, most datasets depend on correlating endoscopic views with pre-operative CT scans or generating sparse point clouds via calibrated stereo-endoscopes. However, the limited availability of training data remains a significant barrier. As a result, existing research of SDE in medical imaging put extra effort into learning through view synthesis \cite{recasens2021,shao2022self,liu2019dense,ozyoruk2021endoslam}, generative models, \cite{tian2024dd,mathew2020augmenting,shen2019context,wei2024absolute} and Shape from Shading \cite{puigvert2023lightdepth} to estimate depth for individual datasets by enhancing unsupervised training efficacy or to overcome illumination changes. With the success of foundation models and their outstanding performance in zero-shot SDE, a few recent works attempted to adapt foundation models trained on natural images to medical imaging \cite{cui2024surgical,cui2024endodac}. However, due to the large domain gap between medical images and natural images, zero-shot generalization of these foundation models is not as good as expected \cite{han2024depth}.

\vspace{0.3cm}
\noindent \textbf{Learning with Noisy Labels.} Noisy labels are a frequent issue in real-world datasets, particularly in medical imaging, where accurate annotations are costly and not always accessible. Despite the substantial performance gains deep neural networks achieve with large volumes of training data, their training processes remain susceptible to overfitting when exposed to noisy labels \cite{zhang2021understanding}. Learning robustly through noisy annotations is thus an important topic. Prior works on robust network learning primarily focus on classification tasks \cite{zhou2020robust,liu2020early,li2023disc}. Fewer SDE studies have focused on dealing with false annotations. Notably, \cite{ranftl2020towards} introduced a trimming strategy for loss calculation. In this approach, outliers are detected and removed by trimming the largest loss residuals in each instance. While this method is effective in many cases, it also removes loss components where the network's estimation significantly deviates from the ground truth, making harder-to-learn cases even more challenging.


In this study, we have curated the largest meta-dataset for endoscopy SDE and trained our EndoOmni model on this extensive mixed dataset. To mitigate label noise, we introduce a self-learning framework that uses both (pseudo) labels and their associated confidence map to guide network training. Additionally, we propose a weighted scale-and-shift-invariant loss to concentrate gradient updates on regions with high-confidence labels.

\setlength{\tabcolsep}{0pt}
\begin{table}[t]
  \centering
  \footnotesize
  \caption{Mixing dataset overview.}
  \begin{threeparttable}
  \resizebox{\columnwidth}{!}{
    \begin{tabular}{cccc}
    \hline
    \textbf{Dataset} & \textbf{Organs} & \textbf{Size} & \textbf{Usage} \\
    \hline
    \multicolumn{4}{c}{\textit{Labeled}} \\
    \hline
    Hamlyn Centre Datasets\textsuperscript{\textdagger} & Kidney & 91,866 fr. & E \\
    C3VD \cite{bobrow2023} & Colon phantom & 10,015 fr. & T \\
    SERV-CT \cite{edwards2020serv} & Liver, kidney, heart & 32 fr. & E \\
    EndoAbS \cite{penza2018endoabs} & Liver, kidney, spleen & 240 fr. & T \\
    SCARED \cite{allan2021stereo} & Abdomen & 17,056 fr. & T \\
    SimColon \cite{rau2019implicit} & Colon simulation & 16,016 fr. & T \\
    EndoMapper \cite{azagra2023endomapper} & Colon simulation & 1,919 fr. & T \\
    Ours-Phantom & Airway phantom & 395 fr. & E \\
    Ours-Patient & Airway & 70,223/5,754 fr. & T/E \\
    \hline
    \multicolumn{4}{c}{\textit{Unlabeled}} \\
    \hline
    EndoSLAM \cite{ozyoruk2021endoslam} & Colon, stomach, etc. & 42,459 fr. & T \\
    EndoMapper \cite{azagra2023endomapper} & Colon & 241,887 fr. & T \\
    ROBUST-MIS \cite{ross2021comparative} & Abdomen & 128,501 fr. & T \\
    EAD \cite{ali2020objective} & Stomach, ureter, etc. & 2,531 fr. & T \\
    Surgical-Vis \cite{zia2021surgical} & - & 3,360 fr. & T \\
    CholecT50 \cite{nwoye2022rendezvous} & Abdomen & 100,863 fr. & T \\
    CVC-ClinicDB \cite{bernal2015wm} & Colon & 612 fr. & T \\
    \hline
  \end{tabular}
  }
  \begin{tablenotes}
    \item Dataset size refers to the number of image frames actually used for training or testing. Original datasets may be selectively extracted to ensure variability. T and E refer to the train and evaluation datasets, respectively. For Ours-Patient, only the training splits are used for training the foundation model, excluding the test splits.
  \end{tablenotes}
  \end{threeparttable}
  \label{tab:datasets}
\end{table}
\let\thefootnote\relax\footnotetext{\textsuperscript{\textdagger} http://hamlyn.doc.ic.ac.uk/vision/}

\begin{figure*}[tbp]
\centerline{\includegraphics[width=0.9\textwidth]{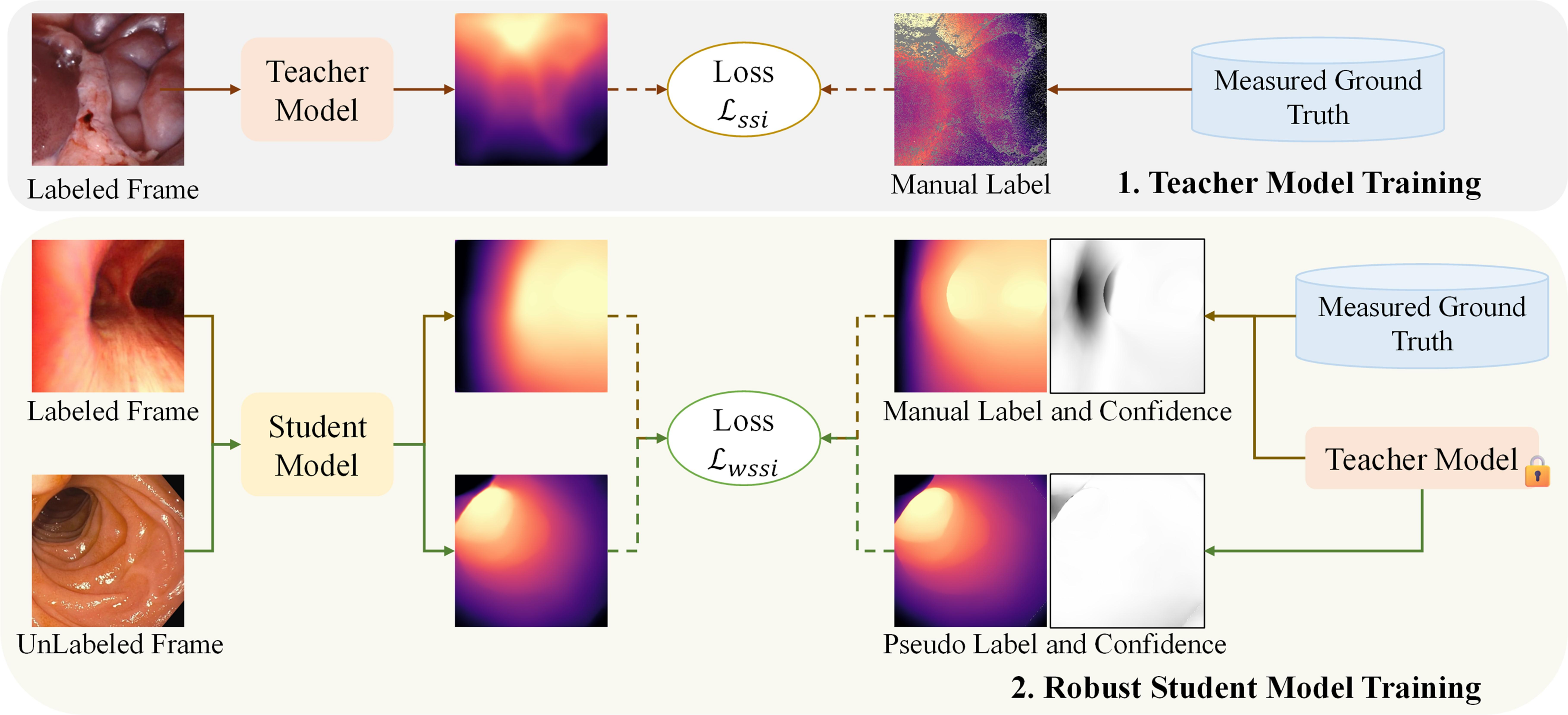}}
\caption{Training framework of EndoOmni: A teacher network is first trained on labeled datasets. Sequentially, a student model is trained on a mix of labeled and unlabeled data, leveraging our robust learning loss guided by the pretrained teacher model. This diverse data and robust training framework enhances the student model's generalization ability.}
\label{fig:diagram}
\end{figure*}

\section{Methodology}
In this study, we establish a robust training framework that utilizes both labeled and unlabeled data for SDE network training. The process starts by training a teacher model $g$ on labeled datasets. Subsequently, a student model $f$ is trained using a combination of labeled data $X^l=\left\{\left(\bm{x}_n^l, \bm{z}_n\right)\right\}_{n=1}^N$ and unlabeled data $X^{u}=\left\{\bm{x}_m^{u}\right\}_{m=1}^M$. The teacher model $g$ supports this process by generating pseudo-labels for the unlabeled data and assessing label confidence for both manually annotated and pseudo labels. The training framework of EndoOmni is illustrated in Fig. \ref{fig:diagram}. 

\subsection{Teacher Model Training}
Following \cite{ranftl2020towards, ranftl2021vision, yang2024depth}, we convert depth labels into disparity space $\bm{d} = \frac{1}{\bm{z}}$ and normalize these within the range 0 to 1 for each depth map. To facilitate training across various datasets without the interference of inconsistent scale and shift factors across datasets, we train the teacher model with scale-and-shift invariant (SSI) loss \cite{ranftl2020towards}:

\begin{equation}
L_{\text{ssi}} = \frac{1}{HW} \sum_{i=1}^{HW} \rho(\hat{\bm{d}}_i, \bm{d}_i),
\end{equation}

\noindent where $\hat{\bm{d}} = g(x)$ and $\bm{d}$ represent the predicted and ground truth depth, respectively. $H$ and $W$ are the image height and width. $\rho$ is formulated as:

\begin{equation}
\rho(\hat{\bm{d}}, \bm{d}) = |\hat{\bm{d}}^* - \bm{d}^*|.
\label{eq:ssi}
\end{equation}

Here, $\hat{\bm{d}}^*$ and $\bm{d}^*$ are calculated by aligning the prediction $\hat{\bm{d}}$ and ground truth $\bm{d}$ to have zero translation and unit scale:

\begin{equation}
\bm{d}^* = \frac{\bm{d} - t(\bm{d})}{s(\bm{d})}, \quad \hat{\bm{d}}^* = \frac{\hat{\bm{d}} - t(\hat{\bm{d}})}{s(\hat{\bm{d}})},
\end{equation}

\noindent where $t(\bm{d})$ is the median of $\bm{d}$, and $s(\bm{d})$ is defined as:

\begin{equation}
s(\bm{d}) = \sum_{i=1}^{HW} |\bm{d}_i - t(\bm{d}_i)|.
\end{equation}

In this way, we decouple the effects of scale and translation differences between datasets, enhancing the model's focus on intrinsic predictive accuracy.

{The teacher model is trained with all labeled data listed in Table \ref{tab:datasets} with notation "T".} Accurately annotating depth labels in endoscopic data presents a significant challenge compared to general open scenes because high-precision sensors are impractical to set up during clinical procedures. Consequently, existing datasets generate annotations using stereo cameras \cite{allan2021stereo, edwards2020serv, penza2018endoabs}, registration to CT scans \cite{bobrow2023}, or synthetic data \cite{azagra2023endomapper, rau2019implicit}. To expand the training data, we have curated an additional dataset for bronchoscopy. Depth annotations are sourced using registration-based methods, airway meshes from patient CT scans are reconstructed, and video frames are manually registered to these in a virtual endoscopic view. Depth labels are generated by rendering depth maps in the airway model with the registered endoscopic poses. As a result, our teacher model leverages these extensive datasets, further enhancing training data diversity by generating pseudo-labels for unlabeled datasets to guide the student model training.

\subsection{Robust Student Model Training}
To extend training data diversity for the student model, besides labeled data, we have incorporated unlabeled endoscopic data from datasets intended for other tasks. All datasets used for training and evaluation of the student model are listed in Table \ref{tab:datasets} with notation "T" and "E". Each dataset introduces unique challenges due to variations in annotation accuracy—depth labels from stereo cameras can be sparse and imprecise due to calibration errors and environmental factors, while labels from registration methods may suffer from errors in registration and CT reconstruction inaccuracies. Leveraging this mix of unlabeled data enhances the breadth of our training data, yet introduces the risk of noisy pseudo-labels from the teacher model, which can mislead the student model's training. This is particularly problematic in medical imaging, where fewer labeled data are available to train the teacher model, leading to its unstable performance on out-of-distribution unlabeled data. Such noise in training could result in degraded generalization performance during practical application. To address this, we propose to mitigate the impact of these inaccuracies by estimating the confidence of each labeled pixel. We then adjust the learning weight for each pixel based on its estimated confidence. This approach focuses gradient updates on regions with higher confidence, while reducing the influence of gradients from noisier regions.

\begin{figure*}[t]
\centerline{\includegraphics[width=\textwidth]{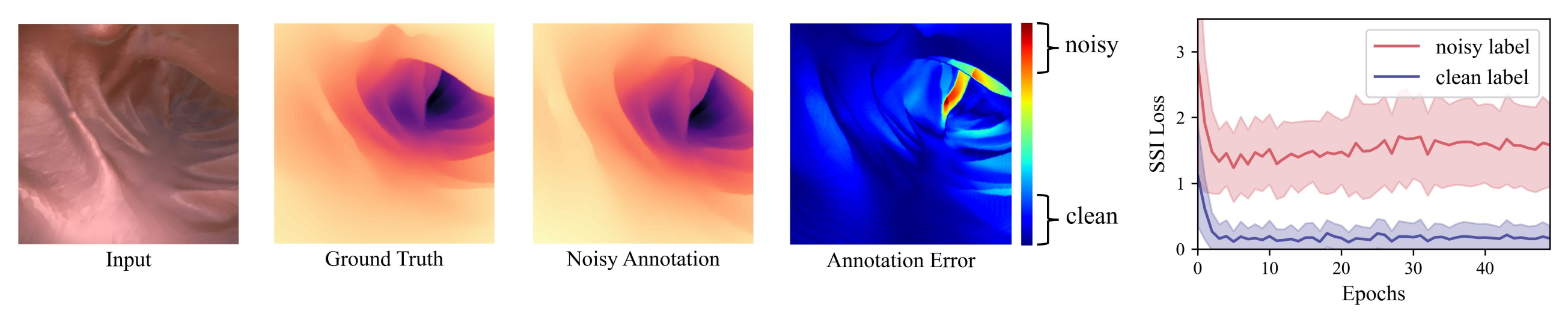}}
\caption{Model learning dynamics. Noisy annotations are created by misaligning image frames with ground truth depth labels. Annotation error is the L1 distance between ground truth and noisy annotations. Clean and noisy pixels are sampled based on annotation error, with loss convergence shown as mean $\pm$ standard deviation. The loss on clean pixels quickly converges, while it remains higher on noisy pixels throughout training.}
\label{fig_labeled}
\end{figure*}

\vspace{0.3cm}
\noindent \textbf{Learning Bias towards Clean Labeled Data.}
Previous studies on learning from noisy labels in classification problems have demonstrated that deep neural networks tend to first fit clean labels before potentially memorizing mislabeled ones \cite{zhou2020robust, liu2020early, li2023disc}. This is because clean labels provide consistent and accurate gradient updates, leading to faster and more effective learning. The learning phenomenon has motivated a deeper examination of the teacher model's training dynamics. Unlike classification tasks, SDE convergence can vary at a pixel level, depending on the precision of the annotations. We experiment with introducing annotation inaccuracies to the synthetic dataset SimColon \cite{rau2019implicit} by misaligning image frames with depth labels, as depicted in Fig. \ref{fig_labeled}. The teacher model rapidly converges for pixels with accurate labels, while those with noisy labels continue to exhibit elevated loss levels. This differential convergence pattern allows us to evaluate the confidence in each labeled pixel by examining their learning statuses as determined by the teacher model. Specifically, if the teacher model demonstrates early convergence at a pixel label, it suggests that the label is likely accurate. On the other hand, persistent non-convergence may signal potential inaccuracies. Based on these insights, we propose estimating the confidence of depth labels at each pixel using the formula:
\begin{equation}
p(\bm{d}_i \mid \bm{x}_i) \sim \rho(g(\bm{x}_i) , \bm{d}_i),
\end{equation}
\noindent where the confidence of depth labels is determined by the SSI discrepancy between manual labels and pretrained teacher model prediction based on Eq. \ref{eq:ssi}.

\begin{figure*}[t]
\centerline{\includegraphics[width=\textwidth]{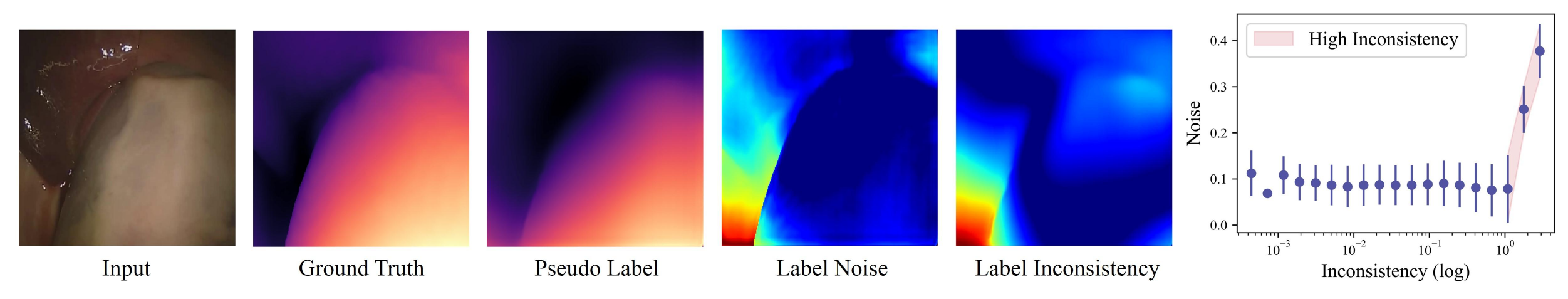}}
\caption{Correlation between pseudo label inconsistency and noise. Pseudo labels are the average predictions from the teacher network across augmentations. Label noise is the SSI difference between ground truth and pseudo labels, while label inconsistency measures the SSI discrepancy between teacher outputs. The binned analysis shows that noise increases significantly in high inconsistency regions (80th percentile threshold = 1.0), indicating that higher inconsistency generally suggests larger pseudo label errors.}
\label{fig_unlabeled}
\end{figure*}

To leverage this insight, we convert the label confidence into a weight map that prioritizes gradient updates from regions with higher confidence during training. This approach minimizes the influence of potentially less accurate regions on the network’s learning process. Accordingly, we define the weighted scale-and-shift invariant loss (WSSI) for labeled data formulated as:

\begin{equation}
L_{\text{wssi}}^{\text{labeled}} = \frac{1}{U} \sum_{i=1}^{HW} \bm{w}_i \cdot \rho(f(\bm{x}_i) , \bm{d}_i),
\end{equation}

\noindent where $U = \sum_{i=1}^{HW} \bm{w}_i$ and the weight matrix $\bm{w}$ is defined as:

\begin{equation}
    \bm{w} = \exp\left(-\frac{\rho(g(\bm{x}) , \bm{d})^2}{\tau_1}\right),
\end{equation}

\noindent with the temperature parameter $\tau_1$ empirically set to 0.5, based on preliminary experiments. This weight matrix applies an exponential decay based on the squared discrepancy between the manual labels and the teacher model’s predictions, ensuring a rapid reduction in the influence of label outliers on the overall training loss, while allowing contributions from labels that are more likely to be accurate.

\vspace{0.3cm}
\noindent \textbf{Learning Bias towards Clean Pseudo-labels.}
Previous research in self-supervised learning for classification has shown that a model's output for a sample is likely to be an accurate pseudo-label if it remains consistent across different augmentations of that sample \cite{zhou2020robust}. We extend this analysis to monocular depth estimation by comparing the zero-shot prediction errors of the teacher model and the inconsistencies in predictions across different data augmentations. As demonstrated in Fig \ref{fig_unlabeled}, the pseudo-label error map closely aligns with the inconsistency map of the teacher model. Based on the observed output patterns, we hypothesize that the confidence of a pseudo-label can be estimated by the consistency of the teacher model's outputs on different augmentations of the input sample. Formally, given an input $x$ and a teacher model $g$, the confidence of a pseudo-label pixel can be expressed as:

\begin{equation}
p(g(\bm{x}_i) \mid \bm{x}_i) \sim \sigma\left(\{g(\hat{\bm{x}}_i^{(k)})\}_{k=1}^K\right),
\end{equation}

\noindent where $\sigma(\cdot)$ measures the consistency of estimates across augmented inputs $\hat{\bm{x}}_i^{(k)}$. For training efficiency, we simplify the confidence measurement by setting the number of augmentations $K=2$, using random flipping and rotation. Consequently, we define the weighted scale-and-shift invariant loss (WSSI) for unlabeled data as:

\begin{equation}
L_{\text{wssi}}^{\text{unlabeled}} = \frac{1}{V} \sum_{i=1}^{HW} \bm{v}_i \cdot \rho(f(\bm{x}_i) - \bar{g}(\bm{x}_i)),
\end{equation}

\noindent where $V=\sum_{i=1}^{HW} \bm{v}_i$. The pseudo label $\bar{g}(\bm{x})$ is computed as $\bar{g}(\bm{x}) = \frac{g(\bm{x}) + g(\hat{\bm{x}})}{2}$, with $\hat{\bm{x}}_i$ representing augmented version of $\bm{x}_i$. The weight matrix $\bm{v}$ is defined as:

\begin{equation}
    \bm{v} = \exp\left(-\frac{\rho(g(\bm{x}) , g(\hat{\bm{x}}))^2}{\tau_2}\right),
\end{equation}

\noindent with the temperature parameter $\tau_2$ empirically set to 0.1 based on preliminary experiments. The weight for unlabeled data emphasizes reliable pseudo labels where the teacher model demonstrates higher consistency across different augmentations, and de-emphasizes contribution from uncertain pseudo labels to avoid disrupting student model training.

Compared to the trimmed scale-and-shift invariant loss (TSSI) \cite{ranftl2020towards}, which relies on the instantaneous outputs of the student network to detect label outliers, our WSSI leverages a pretrained teacher model that has undergone additional training epochs, yielding more reliable error detection. Furthermore, instead of discarding a fixed percentage of data points, our method uses a soft weighting matrix that dynamically adjusts the influence of each data point, thus preserving the original data distribution.





\setlength{\tabcolsep}{1.3pt}
\begin{table}[tbp]
  \centering
  \footnotesize
  \caption{Zero-shot RDE on Hamlyn Dataset.}
  \resizebox{\columnwidth}{!}{
    \begin{tabular}{cccccc}
    \toprule
    \textbf{Method} & \textbf{AbsRel↓} & \textbf{SqRel↓} & \textbf{RMSE↓} & \textbf{RMSElog↓} & $\delta 1$↑ \\
    \hline
    \textbf{EDM} \cite{recasens2021} & 0.185 & 5.424 & 16.100  & 0.255 & 0.732 \\
    \textbf{AF-SfMLearner$^{\ddagger}$} \cite{shao2022self} & 0.168 & 4.440  & 13.870 & 0.204 & 0.770 \\
    \textbf{Surgical-DINO} \cite{cui2024surgical} & 0.146 & 3.216 & 11.974 & 0.178 & 0.801 \\
    \textbf{EndoDAC} \cite{cui2024endodac} & 0.138 & 2.796 & 11.491 & 0.171 & 0.813 \\
    \textbf{IID-SfmLearner} \cite{li2024image} & 0.157 & 2.995 & 11.621 & 0.192 & 0.776 \\
    \textbf{DepthAnything-L} \cite{yang2024depth} & 0.174  & 3.829  & 13.226  & 0.210  & 0.750  \\
    \textbf{DepthAnything2-L} \cite{yang2024depth2} & 0.229  & 5.949  & 16.718  & 0.269  & 0.649  \\
    \textbf{EndoOmni (Ours)} & \textbf{0.125} & \textbf{2.337} & \textbf{10.858} & \textbf{0.160} & \textbf{0.827} \\
    \bottomrule
    \end{tabular}
    }
    \begin{tablenotes} 
\item \textbf{Best} performances are highlighted. All results obtained by recovering scale before evaluation. 
$\ddagger$ denotes in-domain evaluation.
    \end{tablenotes}
  \label{tab:rde_hamlyn}%
\end{table}%

\begin{table}[t]
  \centering
  \footnotesize
  \caption{Zero-shot RDE on SERV-CT.}
    \resizebox{\columnwidth}{!}{
    \begin{tabular}{cccccc}
    \toprule
    \textbf{Method} & \textbf{AbsRel↓} & \textbf{SqRel↓} & \textbf{RMSE↓} & \textbf{RMSElog↓} & $\delta 1$↑ \\
    \midrule
    \textbf{SfMLearner} \cite{zhou2017unsupervised} & 0.151  & 3.917  & 17.451  & 0.191  & 0.779  \\
    \textbf{Fang et al.} \cite{fang2020towards} & 0.149  & 3.099  & 15.564  & 0.188  & 0.787  \\
    \textbf{DeFeat-Net} \cite{spencer2020defeat} & 0.114  & 1.946  & 12.588  & 0.153  & 0.873  \\
    \textbf{SC-SfMLearner} \cite{bian2019unsupervised} & 0.117  & 2.015  & 12.415  & 0.148  & 0.852  \\
    \textbf{Monodepth2} \cite{godard2019digging} & 0.123  & 2.205  & 12.927  & 0.152  & 0.856  \\
    \textbf{Endo-SfM} \cite{ozyoruk2021endoslam} & 0.116  & 2.014  & 12.493  & 0.143  & 0.864  \\
    \textbf{AF-SfMLearner} \cite{shao2022self} & 0.102  & 1.632  & 11.092  & 0.131  & 0.898  \\
    \textbf{IID-SfmLearner} \cite{li2024image} & 0.123 & 1.870 & 10.985 & 0.153 & 0.833 \\
    \textbf{MonoPCC} \cite{wang2024monopcc} & 0.091 & 1.252 & 10.059 & 0.116 & 0.915 \\
    \textbf{EndoDAC}\cite{cui2024endodac} & 0.079 & 0.983 & 8.723 & 0.103 & 0.945 \\
    \textbf{DepthAnything-L} \cite{yang2024depth} & 0.080  & 0.906  & 7.212  & 0.098  & 0.928  \\
    \textbf{DepthAnything2-L} \cite{yang2024depth2} & 0.130  & 2.125  & 10.578  & 0.160  & 0.818  \\
    \textbf{EndoOmni (Ours)} & \textbf{0.053} & \textbf{0.432} & \textbf{5.922} & \textbf{0.070} & \textbf{0.988} \\
   
    \bottomrule
    \end{tabular}%
    }
    \begin{tablenotes} 
\item \textbf{Best} performances are highlighted. All results were obtained by recovering the scale before evaluation. 
    \end{tablenotes}
  \label{tab:rde_servct}%
\end{table}%

\section{Experiments}
We conducted extensive experiments to evaluate the zero-shot relative depth estimation (RDE), fine-tuned to metric depth estimation (MDE), and zero-shot MDE on unseen datasets. We also analyzed the WSSI loss in ablation studies. Furthermore, we explored utilizing our depth estimation foundation model for other related tasks on endoscopic images.

\subsection{Implementation Details}
We adopt the DINOv2 \cite{oquab2024dinov2} encoder and DPT \cite{ranftl2021vision} decoder for both the teacher and student models, as introduced by \cite{yang2024depth}. We first train the teacher model with a ViT-L backbone \cite{dosovitskiy2021an} on all labeled training sets until convergence. We then train the student model with a batch size of 32 until convergence under the teacher model's guidance. We offer three student model scales based on ViT-S, ViT-B, and ViT-L architectures respectively. Following the approach in \cite{yang2024depth}, we challenge the student model with strong perturbations including color jittering and Gaussian blurring. Additionally, a semantic perception technique aligning backbone features with DINOv2 \cite{oquab2024dinov2} is adopted. Labeled data and unlabeled data are mixed in a 1:3 ratio for each input batch. For both teacher and student network training, we initialize the encoder with pretrained weights \cite{yang2024depth} for faster convergence and randomly initialize the decoder weights. We use a learning rate of $5\times10^{-6}$ and a learning rate factor of 0.1:1 for the encoder and decoder. The networks are trained with the Adam optimizer with a linear decay learning rate.

For a fair comparison with recent work in medical image depth estimation, we fine-tune our foundation model $f$ to produce a refined model $\hat{f}$ for zero-shot relative depth estimation (RDE), leaving only an unknown scale factor. Note that the key contribution remains the foundation model $f$. The student model $f$ traversed the SCARED dataset once using scale-invariant loss. SCARED was selected for its high annotation accuracy and sufficient training frames. Formally, the scale-invariant loss\cite{nips2014_7bccfde7} is defined as:

\begin{equation}
L_{\text{si}} = \sqrt{\frac{1}{HW} \sum_{i=1}^{HW} (h_i)^2 - \frac{\lambda}{(HW)^2} \left(\sum_{i=1}^{HW} h_i\right)^2},
\end{equation}

\noindent where $h_i^k = \log \dot{f}(\bm{x}_i)_i^k - \log \bm{y}_i^k$ is the value of the log-depth difference map at position $i$ and scale $k$.

During experiments, we included our scale-invariant version of model $\hat{f}$ with ViT-B backbone, which brings the parameter count closer to that of previous methods. We used the same median scaling preprocessing as existing methods on SDE in medical imaging, to ensure a fair comparison.

\begin{table}[t]
  \centering
  \footnotesize
  \caption{Fine-tuned to MDE on Hamlyn Dataset.}
    \resizebox{\columnwidth}{!}{
    \begin{tabular}{cccccc}
    \toprule
    \textbf{Method} & \textbf{AbsRel↓} & \textbf{SqRel↓} & \textbf{RMSE↓} & \textbf{RMSElog↓} & $\delta 1$↑ \\
    \midrule
    \textbf{EDM-S} \cite{recasens2021} & 0.235  & 7.045  & 17.500  & 0.253  & 0.671  \\
    \textbf{EDM-MS} \cite{recasens2021} & 0.327  & 10.693  & 25.009  & 0.440  & 0.337  \\
    \midrule
    \textbf{DepthAnything} \cite{yang2024depth} & 0.135  & 2.264  & 10.046  & 0.167  & 0.810  \\
    \textbf{EndoOmni (Ours)} & \textbf{0.127} & \textbf{1.993} & \textbf{9.169} & \textbf{0.144} & \textbf{0.865} \\
    \bottomrule
    \end{tabular}%
    }
    \begin{tablenotes} 
 \item EDM-S and EDM-MS correspond to Endo-Depth-and-Motion Stereo and Mono+Stereo respectively.
    \end{tablenotes}
  \label{tab:mde_hamlyn}%
\end{table}%

\begin{table}[t]
  \centering
  \footnotesize
  \caption{Fine-tuned to MDE on patient bronchoscopy dataset.}
    \resizebox{\columnwidth}{!}{
    \begin{tabular}{cccccc}
    \toprule
    \textbf{Method} & \textbf{AbsRel↓} & \textbf{SqRel↓} & \textbf{RMSE↓} & \textbf{RMSElog↓} & $\delta 1$↑ \\
    \midrule
    \textbf{Cycle-GAN} \cite{zhu2017unpaired} & 1.153 & 18.594 & 13.26 & 0.736 & 0.185 \\
    \textbf{DD-VNB} \cite{tian2024dd} & 0.656 & 5.443 & 7.631 & 0.554 & 0.320 \\
    \midrule
    \textbf{DepthAnything} \cite{yang2024depth} & 0.404 & 1.270  & 2.761 & 0.364 & 0.482 \\
    \textbf{EndoOmni (Ours)} & \textbf{0.345} & \textbf{0.987} & \textbf{2.568} & \textbf{0.337} & \textbf{0.519} \\
    \bottomrule
    \end{tabular}%
    }
  \label{tab:mde_patient}%
\end{table}%

\begin{figure*}[t]
\centerline{\includegraphics[width=\textwidth]{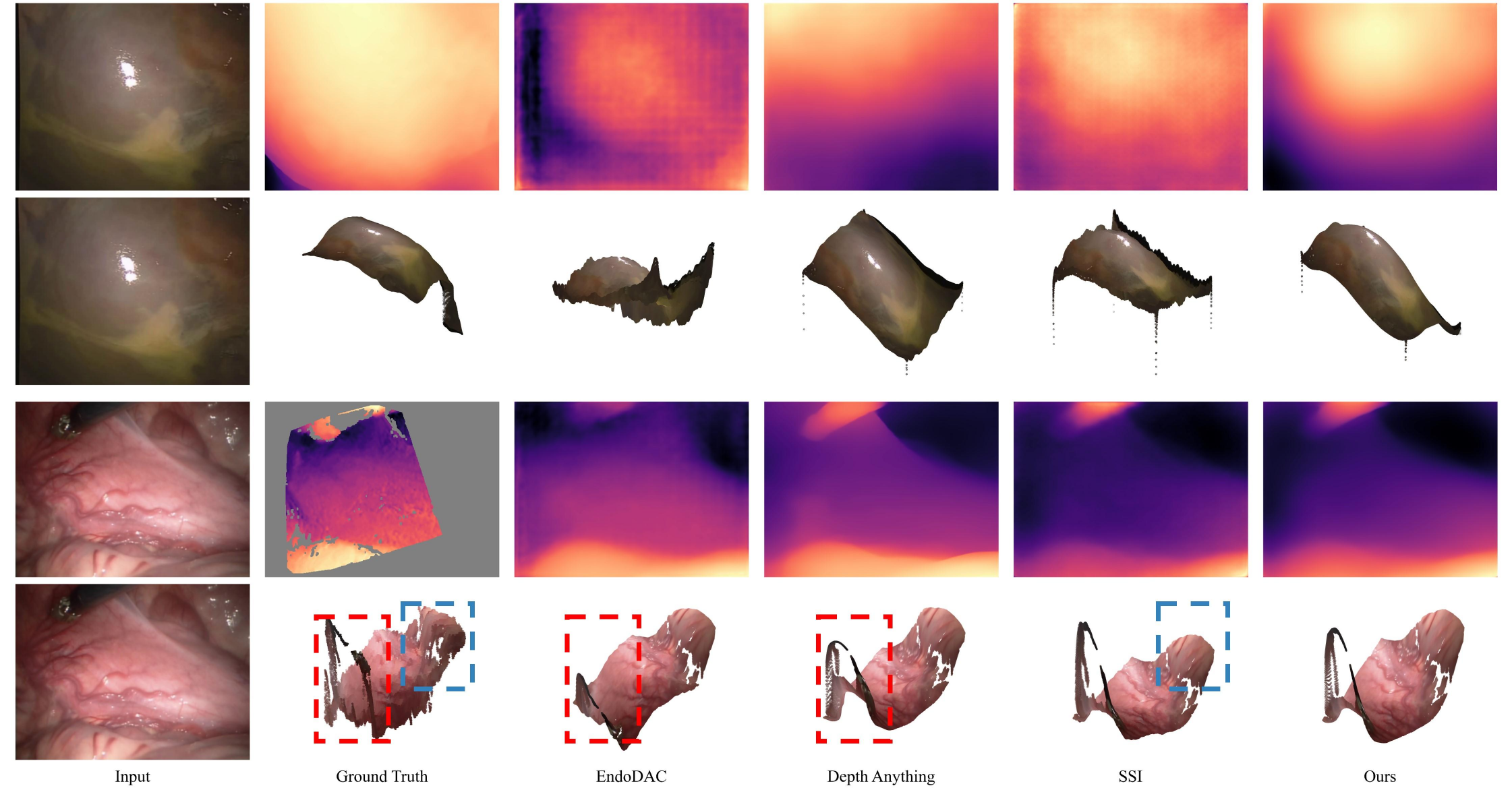}}
\caption{Zero-shot performance of EndoOmni on SERV-CT (top two rows) and the Hamlyn Dataset (bottom two rows), compared with EndoDAC \cite{cui2024endodac}, the leading SOTA method for endoscopy, and Depth Anything \cite{yang2024depth}, the top-performing SOTA foundation model. Quantitative results are also provided for the model without our robust training loss, denoted as SSI. We show corresponding point clouds rendered from the same viewpoint. Misalignments with ground truth are highlighted in boxes on the bottom row for easy identification.}
\label{fig:quantitative}
\end{figure*}

\begin{table}[t]
  \centering
  \footnotesize
  \caption{Zero-shot MDE on phantom bronchoscopy dataset.}
    \resizebox{\columnwidth}{!}{
    \begin{tabular}{cccccc}
    \toprule
    \textbf{Method} & \textbf{AbsRel↓} & \textbf{SqRel↓} & \textbf{RMSE↓} & \textbf{RMSElog↓} & $\delta 1$↑ \\
    \midrule
    \textbf{Cycle-GAN}  \cite{zhu2017unpaired} & 0.583 & 6.427 & 9.953 & 0.488 & 0.234  \\
    \textbf{DD-VNB} \cite{tian2024dd} & 0.498 & 4.064 & 7.628 & 0.424 & 0.331 \\
    \midrule
    \textbf{Depth Anything} \cite{yang2024depth} & 0.508  & 4.280  & 9.430  & 0.576  & 0.329  \\
    \textbf{EndoOmni (Ours)} & \textbf{0.364} & \textbf{1.914} & \textbf{5.607} & \textbf{0.411} & \textbf{0.382} \\
    \bottomrule
    \end{tabular}%
    }
    \begin{tablenotes} 
\item All methods trained on patient data and tested on phantom data.
    \end{tablenotes}
  \label{tab:mde_phantom}%
\end{table}%

\subsection{Evaluation Metrics}

We assess our approach using five widely recognized metrics of depth estimation: AbsRel (absolute relative error), SqRel (square relative error), RMSE (root mean square error), RMSElog (root mean square logarithmic error), and $\delta_1$, following established works \cite{ranftl2020towards,yang2024depth,recasens2021,cui2024endodac}.

To align with existing research, we perform different preprocessing of depth estimation results for different experiments. We evaluate the foundation model $f$ with per-frame scale $s$ and shift $t$ alignment with ground truth following \cite{ranftl2020towards, ranftl2021vision, yang2024depth} by solving:

\begin{equation}
(s, t)\leftarrow\arg \min _{s, t} \sum_{i=1}^HW\left(s \bm{d}_i+t-\bm{d}_i^*\right)^2.
\end{equation}

For RDE comparison with existing medical imaging research, we perform median scaling \cite{zhou2017unsupervised} for the fine-tuned version model $\hat{f}$ expressed by:

\begin{equation}
\tilde{\bm{d}} = \bm{d}^* \times \frac{\text{median}(\bm{d})}{\text{median}(\bm{d}^*)}.
\end{equation}

For MDE estimation, we do not align depth prediction by any means. We note the preprocessing method used in the result tables.

\subsection{Zero-shot Relative Depth Estimation}
Compared with existing research on SDE in medical imaging, one important advantage of leveraging a foundation model trained on abundant mixed data is accurate depth estimation on unseen datasets during training. Therefore, we validate zero-shot RDE performance on two endoscopic datasets: Hamlyn and SERV-CT \cite{edwards2020serv}. We follow existing works on MDE in medical imaging, fine-tuning the model on the SCARED dataset \cite{allan2021stereo} at the third stage of training of our foundation model and use the same split scheme as existing works \cite{recasens2021,cui2024surgical,cui2024endodac}. As shown in Table \ref{tab:rde_hamlyn} and Table \ref{tab:rde_servct}, our model surpasses existing methods by a large margin. Notably, the best performing existing state-of-the-art (SOTA) method on the Hamlyn and SERV-CT Dataset, EndoDAC \cite{cui2024endodac}, focuses on efficiently adapting Depth Anything-B \cite{yang2024depth} to endoscopic scenes through parameter-efficient fine-tuning. Compared with EndoDAC, our EndoOmni-B improves on all evaluated metrics by a large margin. 
These significant improvements highlight the advantages of robust learning with large, diverse data, which supports training larger models and learning more generalizable features. Our model thus gains a specialized feature understanding of medical data through our curated endoscopy datasets and advanced robust training framework. Quantitative results are shown in Fig. \ref{fig:quantitative}. Notably, we observed that Depth Anything and its derivative, EndoDAC, perform poorly on textureless images with smooth structures, as opposed to images with complex geometry. This could be attributed to the scarcity of such scenes in the training data for general depth estimation tasks, despite their prevalence in endoscopic images. In contrast, our EndoOmni consistently delivers the highest-quality outputs.


\begin{figure*}[tp]
    \centering
    \begin{minipage}[b]{0.45\textwidth}
        \includegraphics[width=\textwidth]{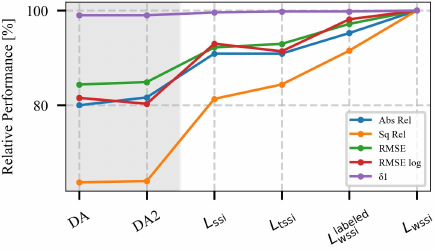}
        \caption*{\footnotesize (a)}
    \end{minipage}
    \hspace{-0pt} 
    \begin{minipage}[b]{0.45\textwidth}
        \includegraphics[width=\textwidth]{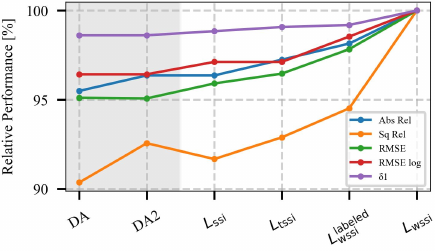}
        \caption*{\footnotesize (b)}
    \end{minipage}
    \caption{Relative performance on (a) SERV-CT and (b) Hamlyn datasets, using our $L_{\text{wssi}}$ as the reference. The four losses in the white region consistently outperform the existing foundation models, Depth Anything (DA) and Depth Anything v2 (DA2), shown in the gray region. Our complete $L_{\text{wssi}}$ achieves superior performance across all metrics, surpassing both existing methods and other loss functions, owing to its robustness against label noise.}

    \label{fig: ablation on losses}
\end{figure*}

\begin{table*}[t]
  \centering
  \footnotesize
  \caption{{Ablation on the size of network backbone.}}
    \begin{tabular}{c|ccccc|ccccc}
    \hline
    \multirow{2}[2]{*}{\textbf{Backbone}} & \multicolumn{5}{c|}{\textbf{SERV-CT}} & \multicolumn{5}{c}{\textbf{Hamlyn}} \\
          & \textbf{AbsRel↓} & \textbf{SqRel↓} & \textbf{RMSE↓} & \textbf{RMSElog↓} & \textbf{$\delta$1↑} & \textbf{AbsRel↓} & \textbf{SqRel↓} & \textbf{RMSE↓} & \textbf{RMSElog↓} & \multicolumn{1}{c}{\textbf{$\delta$1↑}} \\
    \hline
    \textbf{ViT-S} & 0.048 & 0.357 & 5.397 & 0.061 & \underline{0.995} & 0.110 & \underline{1.827} & 9.843 & 0.141 & 0.855 \\
    \textbf{ViT-B} & \underline{0.044} & \underline{0.311} & \underline{4.982} & \underline{0.058} & 0.994 & \textbf{0.104} & \textbf{1.653} & \textbf{9.325} & \textbf{0.134} & \textbf{0.872} \\
    \textbf{ViT-L} & \textbf{0.040} & \textbf{0.270} & \textbf{4.716} & \textbf{0.053} & \textbf{0.997} & \underline{0.106} & 1.830 & \underline{9.362} & \underline{0.135} & \underline{0.864} \\
    \hline
    \end{tabular}%
    \vspace{2mm}
    \begin{minipage}{\textwidth}
\centering
\footnotesize 
\textbf{Best} and \underline{second best} performances are highlighted. Recovered scale and shift for all results.
    \end{minipage}
  \label{tab:ablation_size}%
\end{table*}%

\setlength{\tabcolsep}{5pt}
\begin{table}[t]
  \centering
  \footnotesize
  \caption{Fine-tuned to segmentation on Kvasir dataset.}
  \resizebox{\columnwidth}{!}{
    \begin{tabular}{c|ccccc}
    \hline
    \textbf{Methods} & \textbf{Mean Dice} & \textbf{mIoU} & \textbf{Recall} & \textbf{Precision} & \textbf{F2} \\
    \hline
    \textbf{TransNetR} \cite{jha2024transnetr} & 0.871 & 0.802 & 0.884 & 0.907 & 0.874 \\
    \textbf{PraNet} \cite{fan2020pranet} & 0.898 & 0.849 & 0.906 & 0.913 & 0.898 \\
    \textbf{UACANet-S} \cite{kim2021uacanet} & \underline{0.905} & 0.852	 & -& -& - \\
    \textbf{TGA-Net} \cite{tomar2022tganet} & 0.898 & 0.833 & \textbf{0.913} & 0.912 & \underline{0.903} \\
    \textbf{Polyp-SAM-L} \cite{li2024polyp} & 0.901 & \textbf{0.864} & - & - & - \\
    \hline
    \textbf{DINOv2} \cite{oquab2024dinov2} & 0.883 & 0.814 & 0.864 & \textbf{0.941} & 0.883 \\
    \textbf{EndoOmni (Ours)} & \textbf{0.909} & \underline{0.857} & \underline{0.912} & \underline{0.939} & \textbf{0.909} \\
    \hline
    \end{tabular}%
    }
  \label{tab:seg}%
\end{table}%

\setlength{\tabcolsep}{1.5pt}
\begin{table*}[t]
  \centering
  \footnotesize
  \caption{Branch-level endoscope localization accuracy by fine-tuning to airway lumen detection.}
    \begin{tabular}{ccccccccccccc}
    \toprule
    \textbf{Methods} & \textbf{Case1*} & \textbf{Case2} & \textbf{Case3} & \textbf{Case4*} & \textbf{Case5} & \textbf{Case6} & \textbf{Case7*} & \textbf{Case8} & \textbf{Case9*} & \textbf{Case10} & \textbf{Case11} & \textbf{Mean Acc} \\
    \midrule
    \textbf{AirwayNet} \cite{sganga2019autonomous} & 64.33\% & 40.31\% & 26.92\% & 28.95\% & 46.36\% & 63.16\% & 30.89\% & 23.59\% & 38.11\% & 47.60\% & 15.29\% & 36.01\% \\
    \textbf{BronchoTrack} \cite{tian2024bronchotrack} & 65.61\% & 24.26\% & \textbf{91.15\%} & 44.87\% & 59.55\% & 60.77\% & \underline{79.15\%} & 10.77\% & 75.90\% & 74.50\% & \textbf{40.00\%} & 51.68\% \\
    \midrule
    \textbf{DINOv2} \cite{oquab2024dinov2} & \underline{66.24\%} & \underline{70.96\%} & 74.62\% & \underline{66.58\%} & \underline{81.36\%} & \underline{88.04\%} & 73.75\% & \underline{82.05\%} & \textbf{86.97\%} & \underline{76.20\%} & 30.98\% & \underline{69.65\%} \\
    \textbf{EndoOmni (Ours)} & \textbf{75.16\%} & \textbf{72.95\%} & \underline{79.23\%} & \textbf{72.11\%} & \textbf{90.00\%} & \textbf{98.09\%} & \textbf{85.33\%} & \textbf{87.18\%} & \underline{86.32\%} & \textbf{82.66\%} & \underline{35.88\%} & \textbf{75.04\%} \\
    \bottomrule
    \end{tabular}%
    \vspace{2mm} 
    \begin{minipage}{\textwidth}
\centering
\footnotesize 
* indicates the training cases for BronchoTrack. AirwayNet is trained using case-wise synthetic endoscopic frames. \textbf{Best} and \underline{second best} are highlighted.
    \end{minipage}
  \label{tab:detection}%
\end{table*}%

\subsection{Fine-tuned Metric Depth Estimation}
Estimating depth with an absolute scale is critical for downstream tasks such as localization and mapping. However, only a limited amount of existing work in medical imaging focuses on predicting metric depth from monocular endoscopic images, likely due to the relatively small size of each individual dataset. Thus, we aim to provide a powerful foundation model as a promising initialization for fine-tuning MDE using a limited amount of data. Specifically, we initialize the encoder weight with our pretrained student model $f$ and randomly initialize the decoder. The network is fine-tuned on a single dataset following the training strategy by \cite{bhat2023zoedepth}. We used EndoOmni with the ViT-L backbone for the fine-tuning to metric depth experiments, following \cite{yang2024depth}.

\vspace{0.3cm}
\noindent \textbf{In-Domain Metric Depth Estimation.}
We first evaluate in-domain MDE performance, where our model is fine-tuned and evaluated on the same datasets. We utilize the Hamlyn dataset and our bronchoscopy dataset for this experiment. For Hamlyn, we follow the split in \cite{recasens2021}, using sequences 1, 4, 19, and 20 for testing, and the rest for training. For the bronchoscopy dataset, we train and test our model on data recorded during regular inspection procedures for patients. Results are shown in Table \ref{tab:mde_hamlyn} and Table \ref{tab:mde_patient}. Our fine-tuned model surpasses existing methods on all evaluated metrics by a significant margin. By achieving state-of-the-art performance with fine-tuning for only a few epochs, our foundation model shows a generalized feature understanding ability that transfers to MDE on unseen datasets during training.

\vspace{0.3cm}
\noindent \textbf{Zero-Shot Metric Depth Estimation.}
We further evaluate the zero-shot MDE performance of our model on phantom bronchoscopy data. We choose the model fine-tuned on our patient bronchoscopy data for their similar depth range, making it possible for zero-shot generalization. The evaluation test set of phantom data aligns with data used in \cite{tian2024dd}. As shown in Table \ref{tab:mde_phantom}, our model achieves better zero-shot MDE performance compared to SOTA methods on all evaluated metrics with a large margin. Furthermore, we compare using Depth Anything \cite{yang2024depth} pretrained weights to initialize the MDE model encoder and perform zero-shot MDE on phantom bronchoscopy data in Table \ref{tab:mde_phantom}. By training with our robust loss on abundant mixed endoscopic data, our model learns better endoscopy image representations and achieves better zero-shot MDE performance on all evaluated metrics.

\subsection{Ablation Studies}

\noindent \textbf{Robust Training Loss}.
We validate the effectiveness of our proposed robust training loss by analyzing the zero-shot RDE performance following the first two stages of training. Since the second stage employs a scale-and-shift invariant loss, we evaluate RDE by aligning the per-frame scale and shift with ground truth. We compare the performance of four different training losses: (1) \textbf{SSI}, the standard scale-and-shift invariant loss initially proposed by MiDaS \cite{ranftl2020towards}; (2) \textbf{TSSI}, a loss proposed by \cite{ranftl2020towards}, where outliers are identified and removed by trimming the largest loss residuals per instance; (3) \textbf{Ours: WSSI-labeled}, which incorporates the teacher model to assess ground truth confidence, enhancing learning bias for labeled data; and (4) \textbf{Ours: WSSI}, which is our complete weighted scale and shift invariant loss that extends confidence analysis to both labeled and unlabeled data. We evaluate these losses on the SERV-CT and Hamlyn datasets. As shown in Fig. \ref{fig: ablation on losses}, our WSSI consistently outperforms other methods across all metrics on both datasets, demonstrating the robustness of WSSI in handling diverse data distributions. The improvement is particularly pronounced for Sq Rel, which is more sensitive to large prediction errors in the smaller and more challenging regions. While TSSI shows improvements over SSI, the performance gain is marginal when compared to WSSI. This reinforces the idea that trimming loss outliers is not the most efficient approach for improving performance on medical image datasets.


\vspace{0.3cm}
\noindent \textbf{Comparison with Depth Anything}.
Since EndoOmni initializes encoder weights from the general foundation model Depth Anything, we compare our zero-shot RDE performance to demonstrate the improvements in endoscopy data. We also include a comparison with the latest Depth Anything v2 for reference. As shown in Fig. \ref{fig: ablation on losses}, training on a large medical dataset generally improves SDE performance across all metrics, regardless of the evaluated loss function. Notably, our EndoOmni, leveraging the WSSI loss, consistently outperforms all baselines across the evaluated metrics, particularly on the SERV-CT dataset, which offers more accurate dense depth labels, highlighting our model's superior performance.

\vspace{0.3cm}
\noindent \textbf{Ablation on Backbone Size.}
{To accommodate varying application requirements, we provide three student model sizes based on the ViT-S (24.8M parameters), ViT-B (97.5M parameters), and ViT-L (335.3M parameters) architectures. Their zero-shot performance results are presented in Table \ref{tab:ablation_size}. The results indicate that EndoOmni’s performance shows only a slight decline as the backbone parameter count decreases, making it suitable for resource-constrained environments.}

\subsection{Fine-tuned to Polyp Segmentation}
Although originally trained for the SDE task, EndoOmni is expected to transfer effectively to other downstream tasks in endoscopy. To validate this capability, we applied EndoOmni with the ViT-L backbone to the task of polyp segmentation. We followed the data split of the established benchmark \cite{fan2020pranet} and conducted experiments on the Kvasir-SEG dataset \cite{jha2020kvasir}. For this task, we modified EndoOmni by removing the final ReLU activation and adding batch normalization and dropout layers, without introducing any task-specific modules for segmentation. The results, shown in Table \ref{tab:seg}, demonstrate that EndoOmni achieves superior performance compared to existing methods in medical imaging segmentation, highlighting its strong semantic representation capabilities. Notably, we observed a significant improvement in Dice score from 0.883 to 0.909. Additionally, our EndoOmni adaptation for the segmentation task achieved compatible performance with Polyp-SAM-L \cite{li2024polyp}, a fine-tuned Segment Anything (SAM) \cite{kirillov2023segment} foundation model. This observation underscores EndoOmni's potential as a versatile foundation model for multi-task applications in endoscopy.

\subsection{Fine-tuned to Bronchoscopy Localization}
To further evaluate EndoOmni's transferability to other endoscopic tasks, we fine-tuned it to airway lumen detection. This problem was initially introduced by \cite{sganga2019autonomous} and later used for branch-level endoscope localization in the airway \cite{tian2024bronchotrack}. We hypothesize that the geometric structure semantics learned during SDE training enable EndoOmni to transfer learning to anatomical understanding effectively. We initialized the pretrained EndoOmni encoder with the ViT-L backbone and employed a 3-layer MLP to detect airway branches in each frame. The final endoscope localization accuracy was evaluated using the same test set from \cite{tian2024bronchotrack}. As shown in Table \ref{tab:detection}, the fine-tuned EndoOmni outperformed existing specialized methods for branch-level endoscope localization. Additionally, we tested initializing the encoder with the powerful DINOv2 \cite{oquab2024dinov2} pretrained weights, and EndoOmni still outperformed with a comfortable margin. This further indicates that EndoOmni could serve as a robust foundation for other downstream tasks in endoscopy.

\section{Discussion}
Effective generalization to new scenes is crucial for depth estimation methods in real-world applications, especially in medical imaging, where data collection and model fine-tuning are particularly challenging in clinical settings. However, most existing research in medical depth estimation has focused on specific datasets \cite{recasens2021,shao2022self,ozyoruk2021endoslam,puigvert2023lightdepth,li2024image,wang2024monopcc,tian2024dd,mathew2020augmenting,shen2019context,wei2024absolute}.

In contrast, foundation models in general computer vision excel in zero-shot SDE and offer robust generalization across diverse scenes. However, models trained on general scenes suffer considerable performance drops when applied to specialized domains like endoscopic imaging \cite{cui2024endodac}. Consequently, recent efforts have adapted foundation models to medical datasets through parameter-efficient fine-tuning \cite{cui2024surgical,cui2024endodac}, with some focusing on in-domain testing \cite{li2024advancing}. These adaptations are typically a necessary compromise due to lacking an off-the-shelf solution for depth estimation in medical contexts. However, this fine-tuning can lead to representational collapse, which may reduce the foundation models' generalization capabilities across broader domains \cite{aghajanyan2020better}.

EndoOmni directly addresses this gap by offering a robust, off-the-shelf solution for endoscopic SDE that performs well even without fine-tuning. Our approach builds on the straightforward yet effective method of creating a comprehensive endoscopic meta-dataset to train our foundation model. By leveraging the largest meta-dataset for medical SDE, sourced from both public and proprietary datasets, and applying the teacher-student paradigm \cite{hinton2015distilling}, we maximize the use of unlabeled data. Ablation studies (SSI results) show that this approach yields substantial improvements over general-domain foundation models.

Another key contribution of this work is addressing the disturbance of noisy depth labels on medical data, resulting in depth predictions often exhibiting blurry edges (see Fig. \ref{fig:quantitative}, SSI column). In contrast, general-domain foundation models \cite{yang2024depth} produce sharper edges (see Fig. \ref{fig:quantitative}, Depth Anything column). To tackle this, we developed a refined self-learning framework with a robust WSSI loss. Our analysis shows that clean labels converge more reliably, and pseudo-labels become more dependable when the teacher model’s predictions are consistent across augmentations. Building on this, we estimate per-pixel confidence from the teacher model’s behavior, guiding student model training with both labels and confidence. The WSSI loss dynamically adjusts learning weights based on confidence, mitigating noise while preserving challenging regions, ultimately enhancing depth estimation performance.

\section{Conclusion}
In this paper, we introduce EndoOmni, a depth estimation foundation model designed for endoscopy images. Leveraging large, diverse medical datasets and noise-robust loss functions, EndoOmni achieves state-of-the-art zero-shot performance, outperforming both existing methods specific to medical imaging and general-purpose foundation models. When fine-tuned on target datasets for metric depth estimation, the model serves as a powerful initializer, consistently delivering superior performance across both in-domain and out-of-domain scenarios. Moreover, EndoOmni demonstrates significant transferability to other tasks in the realm of endoscopy, surpassing DINOv2 by a considerable margin in polyp segmentation and bronchoscopy localization.

\section*{Acknowledgments}
This work was supported by the Centre of AI and Robotics, Hong Kong Institute of Science and Innovation, Chinese Academy of Sciences, sponsored by InnoHK Funding, HKSAR, and partially supported by Institute of Automation, Chinese Academy of Sciences.










\AtNextBibliography{\small}
\printbibliography
\end{document}